\documentclass{sig-alternate-05-2015}

\usepackage{array}
\usepackage{multirow}
\usepackage{graphicx}
\usepackage{amsmath}
\usepackage{balance}
\usepackage{enumitem}
\usepackage{subfigure}

\setlist{noitemsep}

\usepackage{lipsum}

\usepackage{etoolbox}
\makeatletter
\patchcmd{\maketitle}{\@copyrightspace}{}{}{}
\makeatother

\let\OLDthebibliography\thebibliography
\renewcommand\thebibliography[1]{
  \OLDthebibliography{#1}
  \setlength{\parskip}{0pt}
  \setlength{\itemsep}{0pt plus 0.3ex}
}

\begin{document}



\title{Detecting Sarcasm in Multimodal Social Platforms}

%
%
%
%
%

\numberofauthors{4} 
%
%

\author{
\alignauthor
Rossano Schifanella\\
\affaddr{University of Turin}\\
       \affaddr{Corso Svizzera 185}\\
       \affaddr{10149, Turin, Italy}\\
       \email{schifane@di.unito.it}
\alignauthor
Paloma de Juan\\
       \affaddr{Yahoo}\\
       \affaddr{229 West 43rd Street}\\
       \affaddr{New York, NY 10036}\\
       \email{pdjuan@yahoo-inc.com}
\alignauthor Joel Tetreault\\
       \affaddr{Yahoo}\\
       \affaddr{229 West 43rd Street}\\
       \affaddr{New York, NY 10036}\\
       \email{tetreaul@gmail.com}  
\and
\alignauthor Liangliang Cao\\
       \affaddr{Yahoo}\\
       \affaddr{229 West 43rd Street}\\
       \affaddr{New York, NY 10036}\\
       \email{liangliang@yahoo-inc.com}
}

\maketitle
\begin{abstract}
Sarcasm is a peculiar form of sentiment expression, where the surface sentiment differs from the implied sentiment. The detection of sarcasm in social media platforms has been applied in the past mainly to textual utterances where lexical indicators (such as interjections and intensifiers), linguistic markers, and contextual information (such as user profiles, or past conversations) were used to detect the sarcastic tone. However, modern social media platforms allow to create multimodal messages where audiovisual content is integrated with the text, making the analysis of a mode in isolation partial. In our work, we first study the relationship between the textual and visual aspects in multimodal posts from three major social media platforms, i.e., Instagram, Tumblr and Twitter, and we run a crowdsourcing task to quantify the extent to which images are perceived as necessary by human annotators. Moreover, we propose two different computational frameworks to detect sarcasm that integrate the textual and visual modalities. The first approach exploits visual semantics trained on an external dataset, and concatenates the semantics features with state-of-the-art textual features. The second method adapts a visual neural network initialized with parameters trained on ImageNet to multimodal sarcastic posts. Results show the positive effect of combining modalities for the detection of sarcasm across platforms and methods.   

\end{abstract}

%
%



%
%

%
%

\keywords{Sarcasm; Social Media; Multimodal; Deep Learning; NLP}


\section{Introduction}
\label{sec:introduction}

\begin{figure}[t]
\centering
\includegraphics[width=0.75\columnwidth]{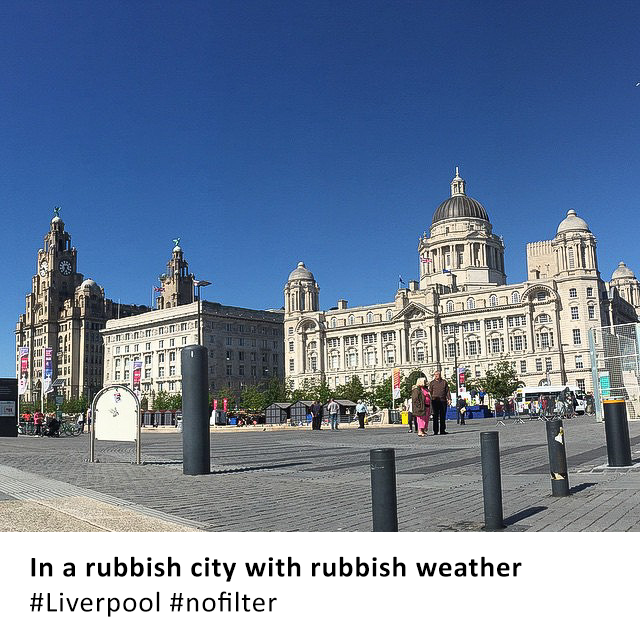}
\caption{Example of an Instagram post where the image is needed to detect the sarcasm. The observation ``rubbish weather'' can only been interpreted correctly by \textit{looking} at the picture. Same holds for ``rubbish city.''}
\label{fig:example-sarcasm}
\end{figure}

Sarcasm is a peculiar form of sentiment expression where the surface sentiment differs from the implied sentiment. Merriam-Webster\footnote{\url{http://www.merriam-webster.com/dictionary/sarcasm}} defines sarcasm as \textit{``the use of words that mean the opposite of what you really want to say especially in order to insult someone, to show irritation, or to be funny.''} Sarcasm is a common phenomenon in social media platforms, and the automatic detection of the implied meaning of a post is a crucial task for a wide range of applications where it is important to assess the speaker's real opinion, e.g., product reviews, forums, or sentiment analysis tools.

Most approaches to sarcasm detection to date have treated the task primarily as a text categorization problem, relying on the insight that sarcastic utterances often contain lexical indicators (such as interjections and intensifiers) and other linguistic markers (such as nonveridicality and hyperbole) that signal the sarcasm. In modern online platforms, hashtags and emojis are common mechanisms to reveal the speaker's true sentiment. These purely text-based approaches have been shown to be fairly accurate across different domains~\cite{Davidov:2010:SRS:1870568.1870582, Gonzalez-Ibanez:2011:IST:2002736.2002850, conf/emnlp/RiloffQSSGH13, conf/emnlp/GhoshGM15, Reyes:2013:MAD:2447287.2447294}.

However, in many occasions this text-only approach fails when contextual knowledge is needed to decode the sarcastic tone. For example, in Figure~\ref{fig:example-sarcasm}, ``rubbish weather'' is the opposite of what the image represents (i.e., beautiful weather). Without this image, the text could be interpreted as a negative comment about the weather in Liverpool. Recently, several approaches~\cite{DBLP:conf/icwsm/BammanS15, Rajadesingan:2015:SDT:2684822.2685316, DBLP:conf/acl/JoshiSB15,khattri-EtAl:2015:WASSA, DBLP:conf/wise/WangWWR15} have integrated contextual cues (e.g., the author's profile, author's past posts and conversations) with the in-post text, showing consistent improvements when detecting sarcasm.

Previous approaches have failed to consider the media linked to the posts as a possible source of contextual information. Tweets, for example, can have audiovisual content attached to the text. Multimodality is the combination of modes of communication (i.e., text, images, animations, sounds, etc.) with the purpose to deliver a message to a particular audience, and it is present in all major social media platforms. 

In this work, we leverage the contextual information carried by visuals to decode the sarcastic tone of multimodal posts. Specifically, we consider two types of visual features with different model fusion methods for sarcasm detection. The first approach exploits visual semantics trained on an external dataset, and concatenates the semantics features with state-of-the-art text features. The second method adapts a visual neural network initialized with parameters trained on ImageNet to multimodal (text+image) sarcastic posts. In both methods, we find that visual features boost the performance of the textual models.

We summarize our main contributions as follows:
\begin{itemize}
  \item We study the interplay between textual and visual content in sarcastic multimodal posts for three main social media platforms, i.e., Instagram, Tumblr and Twitter, and discuss a categorization of the role of images in sarcastic posts. 
  \item We quantitatively show the contribution of visuals in detecting sarcasm through human labeling. This data will be shared with the research community.
  \item We are the first to propose and empirically evaluate two alternative frameworks for sarcasm detection that use both textual and visual features. We show an improvement in performance over textual baselines across platforms and methods. 
\end{itemize}

We first discuss related work in Section~\ref{sec:related}. We then describe our data in Section~\ref{sec:data}, and introduce a categorization of the different roles images can play in a sarcastic post in Section~\ref{sec:characterization}. In the same section, we describe how we collect human judgments to build a gold set, and analyze the distribution of posts with respect to the proposed categories. Section~\ref{sec:methods} describes the details of the two methods for sarcasm detection, and Section~\ref{sec:evaluation} presents the experiments carried out to evaluate the frameworks, and their results. Finally, Section~\ref{sec:conclusions} concludes the paper, and points to future work.
\section{Related Work}\label{sec:related}

\noindent \textbf{Sarcasm as linguistic phenomenon.}
While the use of irony and sarcasm is well studied from its linguistic and psychological aspects~\cite{gibbs2007irony}, automatic recognition of sarcasm has become a widely researched subject in recent years due to its practical implications in social media platforms. Starting from foundational work by Tepperman et al.~\cite{proc_tepperman:yeahrightsarcasm:icslp06} which uses prosodic, spectral (average pitch, pitch slope), and contextual (laughter or response to questions) cues to automatically detect sarcasm in a spoken dialogue, initial approaches mainly addressed linguistic and sentiment features to classify sarcastic utterances. 
Davidov et al.~\cite{Davidov:2010:SRS:1870568.1870582} proposed a semi-supervised approach to classify tweets and Amazon products reviews with the use of syntactic and pattern-based features. Tsur et al.~\cite{Tsur10} focus on product reviews and try to identify sarcastic sentences looking at the patterns of high-frequency and content words. Gonz\'{a}lez-Ib\'{a}\~{n}ez et al.~\cite{Gonzalez-Ibanez:2011:IST:2002736.2002850} study the role of lexical (unigrams and dictionary-based) and pragmatic features such as the presence of positive and negative emoticons and the presence of replies in tweets. Riloff et al.~\cite{conf/emnlp/RiloffQSSGH13} present a bootstrapping algorithm that automatically learns lists of positive sentiment phrases and negative situation phrases from sarcastic tweets. They show that identifying contrasting contexts yields improved recall for sarcasm recognition. More recently, Ghosh et al.~\cite{conf/emnlp/GhoshGM15} propose a reframing of sarcasm detection as a type of word sense disambiguation problem: given an utterance and a target word, identify whether the sense of the target word is literal or sarcastic.

\vspace{0.2cm}
\noindent \textbf{Sarcasm as contextual phenomenon.} Recently it has been observed that sarcasm requires some shared knowledge between the speaker and the audience; it is a profoundly contextual phenomenon~\cite{DBLP:conf/icwsm/BammanS15}. 
Bamman et al.~\cite{DBLP:conf/icwsm/BammanS15} use information about the authors, their relationship to the audience and the immediate communicative context to improve prediction accuracy. Rajadesingan et al.~\cite{Rajadesingan:2015:SDT:2684822.2685316} adopt psychological and behavioral studies on when, why, and how sarcasm is expressed in communicative acts to develop a behavioral model and a set of computational features that merge user's current and past tweets as historical context. Joshi et al.~\cite{DBLP:conf/acl/JoshiSB15} propose a framework based on the linguistic theory of context incongruity and introduce inter-sentential incongruity for sarcasm detection
by considering the previous post in the discussion thread. Khattri et al.~\cite{khattri-EtAl:2015:WASSA} present a quantitative evidence that historical tweets by an author can provide additional context for sarcasm detection. They exploit the author's past sentiment on the entities in a tweet to detect the sarcastic intent. Wang at al.~\cite{DBLP:conf/wise/WangWWR15} focus on message-level sarcasm detection on Twitter using a context-based model that leverages conversations, such as chains of tweets. They introduce a complex classification model that works over an entire tweet sequence and not on one tweet at a time. On the same direction, our work is based on the integration between linguistic and contextual features extracted from the analysis of visuals embedded in multimodal posts.

\vspace{0.2cm}
\noindent \textbf{Sarcasm beyond text.}
Modern social media platforms allow to create multimodal forms of communication where audiovisual content integrates the textual utterance. Previous work~\cite{citeulike:12541369} studied how different types of visuals are used in relation to irony in written discourse, and which pictorial elements contribute to the identification of verbal irony. Most scholars who looked at the relationship between verbal irony and images limited themselves to studying visual markers~\cite{45099478}. 
Usually a visual marker is either used to illustrate the literal meaning, or it may also exhibit incongruence with the literal evaluation of an ironic utterance (incongruence between the literal and intended evaluation). Following Kennedy~\cite{gibbs2008cambridge}, the image itself is usually considered not ironic; however, it may sometimes be important in deciding whether a verbal utterance is ironic or not. According to Verstraten~\cite{narratology}, two types of elements play a role in the process of meaning-giving in the visual domain of static images. These include the \textit{mise en sc\`ene} and \textit{cinematographic} techniques. The mise en sc\`ene is concerned with the question of who and/or what is shown, cinematography deals with the question of how something is shown. Despite the similarities in the intent, our work shows few novel points: first of all, we analyze a large sample of non-curated posts from three different social media platforms, while past work focuses mainly on curated content like advertisements, cartoons, or art. Moreover, to the best of our knowledge, we propose the first computational model that incorporates computer vision techniques to the automatic sarcasm detection pipeline.

\vspace{0.2cm}
\noindent \textbf{Making sense of images.} Recently, a number of research studies were devoted to combine visual and textual information, motivated by the progress of deep learning. Some approaches~\cite{DBLP:journals/corr/KirosSZ14, 41473}
pursue a joint space for visual and semantic embedding, others consider how to generate captions to match the image content~\cite{DBLP:journals/corr/MaoXYWY14a, DBLP:journals/corr/MaLSL15},
or how to capture the sentiment conveyed by an image~\cite{Borth:2013:SLO:2502081.2502268, you2015robust}.
 The most similar approach to our work is that of \cite{shutova-kiela-maillard:2016:N16-1} which investigates the fusion of textual and image information to understand metaphors. A key aspect of our work is that it captures the relation between the visual and the textual dimensions as a whole, e.g., the utterance is not a mere description of an image, while in previous studies text is generally adopted to depict or model the content of an image. 
\section{Data}
\label{sec:data}

\begin{table}[t!]
\centering
\begin{tabular}{|l|c|c|}
\hline
\textbf{Platform} & \textbf{Text} & \textbf{Images} \\
\hline
	\textbf{IG} & Optional (up to 2,200 chars) & 1 \\
	\textbf{TU} (photo) & Optional & 1-10 \\
	\textbf{TU} (text) & Required & 0 or more \\
    \textbf{TW} & Required (up to 140 chars) & 0 or more \\
\hline
\end{tabular}
\caption{Text and image limitations.}
\label{tab:limits}
\end{table}

To investigate the role images play in sarcasm detection, we collect data from three major social platforms that allow to post both text and images, namely Instagram (IG), Tumblr (TU) and Twitter (TW), using their available public APIs. Each of these platforms is originally meant for different purposes regarding the type of media to be shared. Whereas Instagram is an image-centric platform, Twitter is a microblogging network. Tumblr allows users to post different types of content, including ``text'' or ``photo''. Regardless of the post type, images (one or more) can be added to textual posts, and captions can be included in photo posts. The text and image restrictions and limitations for each platform are presented in Table~\ref{tab:limits}.

The three platforms allow users to use hashtags to annotate the content, by embedding them in the text (Instagram, Twitter), or by adding them through a separate field (Tumblr). To collect positive (i.e., sarcastic) examples, we follow a hashtag-based approach by retrieving posts that include the tag \textit{sarcasm} or \textit{sarcastic}. This is a technique extensively used to collect sarcastic examples~\cite{conf/emnlp/GhoshGM15}. Additionally, and for all platforms, we filter out posts that are not in English, and remove retweets (Twitter) and reblogs (Tumblr) to keep the original content only and avoid duplicates.

\begin{table}[t!]
\centering
\begin{tabular}{|l|r|r|r|r|}
\hline
\textbf{Platform} & \multicolumn{1}{c|}{\textbf{\#Posts}} & \multicolumn{1}{c|}{\textbf{w/Text}} & \multicolumn{1}{c|}{\textbf{w/Images}} & \multicolumn{1}{c|}{\textbf{w/Both}} \\
\hline
	\textbf{IG} & 517,229 & 99.74\% & 100\% & 99.74\% \\
	\textbf{TU} & 63,067 & 94.22\% & 45.99\% & 40.22\% \\
    \textbf{TW} & 20,629 & 100\% & 7.56\% & 7.56\% \\
\hline
\end{tabular}
\caption{Presence of textual and visual components.}
\label{tab:textimgdistrib}
\end{table}

Table~\ref{tab:textimgdistrib} shows the distribution of posts with text, image(s), or both for each of the three platforms. Instagram is the platform where the the textual and visual modalities are most used in conjunction; in fact, almost the totality of posts have a caption accompanying the image. In contrast, less than 8\% of the posts on Twitter contain images. Among the 63K Tumblr posts, 56.96\% are of type ``text'', and 43.04\% are of type ``photo''. This means that most of the photo posts contain also text (similar to Instagram, but without the limitation on the number of images), but very few of the text posts contain images (similar to Twitter, but without the character limitation).

\vspace{0.2cm} 
\noindent \textbf{Filtering the data.} 

To clean up the data and build our final dataset we apply a series of four filters commonly used in literature~\cite{Gonzalez-Ibanez:2011:IST:2002736.2002850, Davidov:2010:SRS:1870568.1870582, Rajadesingan:2015:SDT:2684822.2685316}. First, we discard posts that do no contain any images, or whose images are no longer available by the time we collect the data; we then discard posts that contain mentions (@username) or external links (i.e., URLs that do not contain the platform name, or ``t.co'' or ``twimg.com'', in the case of Twitter), as additional information (e.g., conversational history, news story) could be required to understand the context of the message. We also discard posts where \textit{sarcasm} or \textit{sarcastic} is a regular word (not a hashtag), or a hashtag that is part of a sentence (i.e., if it is followed by any regular words), as we are not interested in messages that explicitly address sarcasm (e.g., ``I speak fluent sarcasm.''). Finally, we discard posts that might contain memes or \textit{ecards} (e.g., tag set contains \textit{someecards}), and posts whose text contains less than four regular words.
 
\vspace{0.2cm} 
\noindent \textbf{Final dataset.} We randomly sample 10,000 posts from each platform to build our final dataset. Given the limitations of its public API, and the fact that less than 8\% of the sarcastic posts have both text and images, only 2,005 were available for Twitter. We further clean up the data by removing internal links and the tags that we used to collect the samples (\textit{sarcasm} and \textit{sarcastic}).
These posts are composed of two main aspects: a \textit{textual} and a \textit{visual} component. When we speak about the textual component, we are referring not only to the regular words that form the message, but also to \textit{emojis} and \textit{hashtags} that might be part of that message. These three elements (words, emojis and hashtags) are crucial for the interpretation of the post: while regular words are generally used to present the literal meaning, emojis and hashtags are commonly used to reveal the speaker's intended sentiment~\cite{DBLP:journals/corr/JoshiBC16}, or to share contextual cues with the audience to help decode the sarcasm.

\begin{table}[tbp]
\centering
\begin{tabular}{|l|r|r|r|}
\hline
\textbf{Platform}
& \multicolumn{1}{c|}{\textbf{\#Words}} & \multicolumn{1}{c|}{\textbf{\#Emojis}} & \multicolumn{1}{c|}{\textbf{\#Tags}} \\
\hline
	\textbf{IG} & 10.77 & 0.37 & 7.44 \\
	\textbf{TU} & 24.75 & 0.21 & 7.00 \\
    \textbf{TW} & 9.45 & 0.29 & 1.96 \\
\hline
\end{tabular}
\caption{Average number of words, emojis and tags.}
\label{tab:textstats}
\end{table}

Table~\ref{tab:textstats} shows the average number of regular words, emojis and tags (after having removed \textit{sarcasm}/\textit{sarcastic}) per post. Due to its tight character limitation (which also accounts for the hashtags), Twitter is the platform with the shortest text and the lowest number of tags per post. While Tumblr posts are the longest, the average number of tags is similar to that of Instagram, which has in turn the highest tag-to-word ratio. Indeed, Instagram users seem to express heavily through hashtags, especially compared to Twitter users, whose posts have a similar average word count. Both platforms also have a similar emoji-to-word ratio, which is much lower on Tumblr. The fact that there is a character limitation for both Instagram and Twitter might justify the usage of emojis, which are compact representations of concepts and reactions that would be much more verbose if expressed in words.

Finally, we collect 10,000 negative examples from each platform (2,005 from Twitter, to keep the dataset balanced) by randomly sampling posts that do \textbf{not} contain \textit{sarcasm} or \textit{sarcastic} in either the text or the tag set. These negative posts are subject to the same processing described above, when applicable.
To verify that there are no relevant topical differences between the positive and the negative sets that could correlate with the presence/absence of sarcastic cues, we manually examined a sample of positive and negative posts from each platform. We did not observe such differences; however, we did find some recurring topics in the positive set, such as weather, food, fashion, etc., but these topics were also found in the negative set, only along with non-sarcastic observations (e.g., a picture of a greasy slice of pizza would be captioned as ``healthy'' in the positive set, but as ``unhealthy'' in the negative set). This might indicate that the range of topics in the positive set is more limited, but there is a clear overlap with those in the negative set.
\section{Characterizing the Role of Images in Sarcastic Posts}
\label{sec:characterization}

\begin{figure*}
\centering
\subfigure[]
{\label{fig:text}\includegraphics[width=0.3\linewidth]{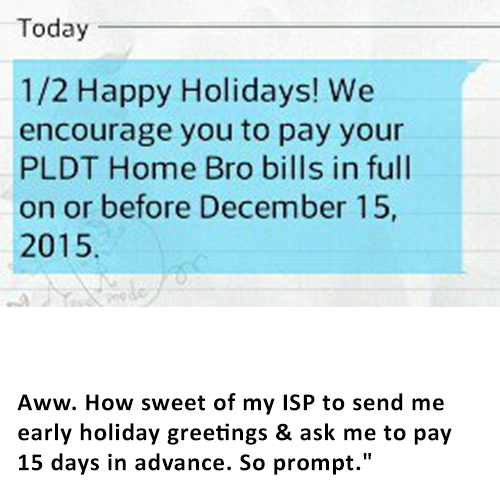}}
\subfigure[]{\label{fig:either}\includegraphics[width=0.3\linewidth]{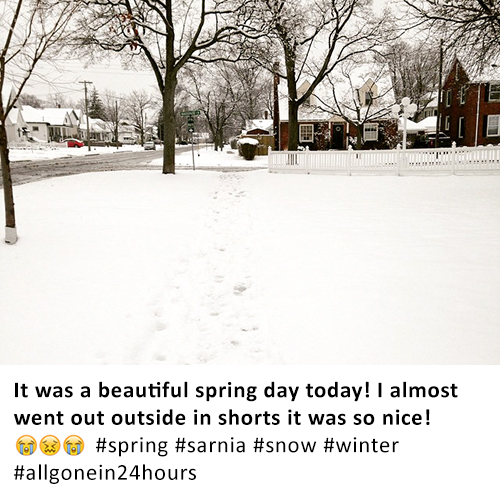}}
\subfigure[]{\label{fig:both}\includegraphics[width=0.3\linewidth]{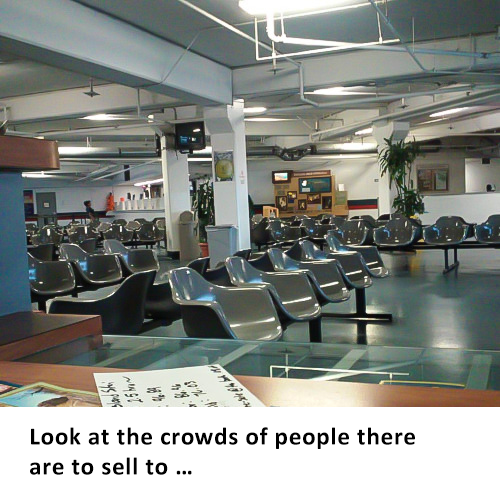}}
\caption{Examples of sarcastic posts.}
\label{fig:examples}
\end{figure*}

As presented in Section~\ref{sec:introduction}, there are two main elements to a sarcastic utterance: the \textit{context} and the \textit{meaning} or \textit{sentiment}. Detecting sarcasm---at a human level---involves evaluating to what extent the intended meaning corresponds to a declared or expected response. If this literal meaning does not agree with the one implied, the utterance will be perceived as sarcastic. In the following sections, we will analyze what role text (i.e., words, emojis and tags) and images play in the conception of sarcasm.

\subsection{Defining a Categorization}
\label{sec:categorization}

To understand what role images play with respect to these two elements, three of the authors independently annotate a set of 100 randomly sampled positive posts from each platform. The question we are looking to answer is: \textit{Is the image \textbf{necessary} to find the post sarcastic?} To answer that, we first identify the posts whose sarcastic nature can be positively determined by just looking at the text. This text, as explained in Section~\ref{sec:data}, can include words, emojis and tags. In many examples, emojis reveal the intended sentiment (in contrast to the literal sentiment presented in the regular text). Hashtags are generally useful to provide context, but can also be used to expose the sentiment. Regardless of whether the sarcastic tone is clear from the text or not, the image can still provide useful clues to understand the intended meaning. The posts where the intended meaning can \textbf{not} be inferred from the text alone are precisely what we are looking for. In these cases, the image turns out to be necessary to interpret the post, providing a depiction of the context, or visual clues to unravel the implied sentiment.

\begin{table}[th]
\centering
\newcolumntype{L}[1]{>{\raggedright\let\newline\\\arraybackslash\hspace{0pt}}m{#1}}
\begin{tabular}{c|c|L{0.38\columnwidth}|L{0.38\columnwidth}|}
\multicolumn{2}{c}{} & \multicolumn{2}{c}{Is the \textbf{TEXT} enough?} \\
\cline{3-4}
\multicolumn{2}{c|}{} & \multicolumn{1}{c|}{Yes} & \multicolumn{1}{c|}{No} \\
\cline{2-4}
	\multirow{2}{*}[2.5em]{\parbox[c]{2mm}{\rotatebox[origin=c]{90}{Does the \textbf{IMAGE} help?}}} &
	\parbox[t]{2mm}{\rotatebox[origin=c]{90}{Yes}} &
    The text is clearly sarcastic; the image provides additional cues for better interpretability and engagement. &
    Both are needed to interpret the post. The clues to understand the intended meaning can be textual or visual. \\
\cline{2-4}
	 &
    \parbox[t]{2mm}{\rotatebox[origin=c]{90}{No}} &
    The text is clearly sarcastic; the image does not provide any added value. &
    Post is \textbf{not} sarcastic. \\
\cline{2-4}
\end{tabular}
\caption{Roles of text and image in sarcastic posts.}
\label{tab:categories}
\end{table}

Table~\ref{tab:categories} summarizes the four possible roles of text and image. We will refer to the category that represents the combination of the two cases to the left as \texttt{Text Only}, as the text from the posts belonging to it should be enough to understand the implied sarcasm. Figures~\ref{fig:text} and \ref{fig:either} are instances of this category. The posts from the top-left case represent a subset of this category, where the image is somewhat redundant, but could replace or augment some of the textual clues. For instance, the image in Figure~\ref{fig:either} would have been necessary if the tags \textit{snow} and \textit{winter} were not part of the text. In this case, also the emojis reveal the implied sentiment, which makes it unnecessary to infer that snow on a spring day is not ``beautiful'' or ``nice'', and that people are not supposed to wear shorts in such weather.

The top right case corresponds to the category that we will call \texttt{Text+Image}, where both modalities are required to understand the intended meaning. Figure~\ref{fig:both} belongs to this category: the image depicts the context that the text refers to. Rather than a sentiment, the text presents an observation (``crowds of people'') that is the opposite of what is shown in the picture (the room is empty). It is worth noting that, regardless of the category, many times the image itself contains text. In this case, the motivation to use an image instead of plain text is generally to provide additional information about the context of this text (e.g., a chat conversation, a screenshot, a street sign, and so on). Figure~\ref{fig:text} is an example of this case.

\subsection{Building a Ground Truth for Sarcasm}
\label{sec:crowdsourcing}

The data collection process described in Section~\ref{sec:data} relies on the ability of the authors to self-annotate their posts as sarcastic using hashtags. Training a sarcasm detector on noisy data is a commonly used approach in literature, especially when that data comes from social media platforms. However, what the audience perceives as sarcastic is not always aligned with the actual intention of the speakers. Our goal is to create a curated dataset of multimodal posts whose sarcastic nature has been agreed on by both the author and the readers, and where both the textual and visual components are required to decode the sarcastic tone. To do that, we use  CrowdFlower,\footnote{\url{http://www.crowdflower.com}} a large crowdsourcing platform that distributes small, discrete tasks to online contributors. The two goals of this annotation task are: 1) characterize the distribution of posts with respect to the categories defined in Section~\ref{sec:categorization}, and 
evaluate the impact of visuals as a source for context for humans; and 2) identify truly sarcastic posts by validating the authors' choice to tag them as such.

\vspace{0.2cm} 
\noindent \textbf{Task interface and setup.}
We focus only on the two main categories of interest, \texttt{Text Only} and \texttt{Text+Image}, and create two independent tasks. In the first task, only the text (including the tags and emojis) is shown to the annotator, along with the question ``Is this text sarcastic?''. The goal is to identify which posts belong to the \texttt{Text Only} category, i.e., posts where the textual component is enough to decode the sarcasm, and the image has a complementary role. We select 1,000 positive posts for this task, using the filters defined in Section~\ref{sec:data}. These posts are randomly sampled from the original sources, with no overlap with the dataset presented in that Section. We collect 5 annotations for each post, where the answer to the question can be ``Yes'' (text is sarcastic), ``No'' (text is \textbf{not} sarcastic) or ``I don't know''.

For the second experiment, we take only those posts that have been marked as non-sarcastic by the majority of the annotators on the first task (i.e., we discard the posts that belong to the \texttt{Text Only} category). Now we present both the textual and visual components, with the question ``Is this post sarcastic?'', and the same possible answers as before. Again, we collect 5 annotations per post. 

The reason we run two independent experiments is to keep the tasks as simple as possible, and to guarantee that the judgment of the annotators is not affected by the knowledge that some information is missing. On the first task, annotators are not aware that the posts originally had one or more images, and are asked to judge them under that impression (same as a text-only based detector would do). If we did a two-step experiment instead, annotators would learn about the missing image(s) after having annotated the very first post, which would invite them to answer ``I don't know'' based on that indication.
We run these experiments for both Instagram and Tumblr. 
Given the limited amount of data that we were able to collect for Twitter, and the fact that only a small percentage of the posts are actually multimodal, we do not build a gold set for this platform.

\vspace{0.2cm} 
\noindent \textbf{Quality control and inter-rater agreement.}
\textit{Test Questions} (also called \textit{Gold Standard} in CrowdFlower jargon) are curated job units that are used to test and track the contributor's performance and filter out bots or unreliable contributors. To access the task, workers are first asked to correctly annotate a set of \textit{Test Questions} in an initial \textit{Quiz Mode} screen, and their performance is tracked throughout the experiment with \textit{Test Questions} randomly inserted in every task, disguised as normal units.

Judgments from contributors whose accuracy on the \textit{Test Questions} is less than 78\% are discarded and marked as not trusted. 

\begin{table}[t!h]
\centering
\begin{tabular}{|l|c|c|c|c|}
\hline
\multirow{2}{*}{\textbf{Task}} & \multicolumn{2}{c|}{\textbf{Matching\%}} & \multicolumn{2}{c|}{\textbf{Fleiss' $\kappa$}} \\
\cline{2-5}
\multicolumn{1}{|c|}{} & \multicolumn{1}{c|}{\textbf{IG}} & \multicolumn{1}{c|}{\textbf{TU}} & \multicolumn{1}{c|}{\textbf{IG}} & \multicolumn{1}{c|}{\textbf{TU}} \\
\hline
	\texttt{Text Only (task 1)} & 80.36 & 76.11 & 0.38 & 0.28 \\
	\texttt{Text+Image (task 2)} & 74.65 & 86.40 & 0.21 & 0.23\\
\hline
\end{tabular}
\caption{Inter-rater agreement.}
\label{tab:agreement}
\end{table}

To assess the quality of the collected data, we measure the level of agreement between annotators (see Table~\ref{tab:agreement}). 
\textit{Matching\%} is the percentage of matching judgments per object. For both experiments, the agreement is solid, with an average value around of 80\%. However, the ratio of matching votes does not capture entirely the extent to which agreement emerges. We therefore compute the standard \textit{Fleiss' $\kappa$}, a statistical measure for assessing the reliability of the agreement between a fixed number of raters. Consistently, the \textit{Fleiss' $\kappa$} shows a Fair level~\cite{Landis77} of agreement where, as expected, the second experiment reaches a lower agreement due to its intrinsic subjectivity and difficulty, even for human annotators~\cite{barbieri-saggion-ronzano:2014:W14-26}.    

\begin{table}[thbp]
\centering
\begin{tabular}{|l|r|r|}
\hline
\multicolumn{1}{|c|}{\textbf{Category}} & \multicolumn{1}{c|}{\textbf{IG}} & \multicolumn{1}{c|}{\textbf{TU}} \\
\hline
	Not sarcastic & 24.8\% & 31.9\% \\
	\texttt{Text Only} & 37.8\% & 23.6\% \\
	\texttt{Text+Image} & \textbf{37.4\%} & \textbf{44.5\%} \\
    	\hspace*{0.5cm} D-80 & 19.1\% & 19.7\% \\
        \hspace*{0.5cm} D-100 & 8.6\% & 14.1\% \\
\hline
\end{tabular}
\caption{Percentage of posts in each category. The D-80 and D-100 subclasses refer to, respectively, posts where at least 80\% or the totality of the annotators agree on the sarcastic nature of the post.}
\label{tab:cfdistrib}
\end{table}

\vspace{0.2cm} 
\noindent \textbf{Results.} Table~\ref{tab:cfdistrib} shows the distribution of the 1,000 posts with respect to the categories described in Section ~\ref{sec:categorization}. For over 60\% of the posts (62.20\% for Instagram, 76.40\% for Tumblr) the text alone (task 1) is \textbf{not} enough to determine whether they are sarcastic or not. However, when those posts are shown with their visual component (task 2), more than half (60.13\% for Instagram, 58.25\% for Tumblr) are actually annotated as sarcastic, i.e., these posts were \textbf{misclassified} as non-sarcastic by the annotators on the first task, so the contribution of the image is crucial. It is interesting to note that a non-negligible fraction of the data (24.80\% for Instagram, 31.90\% for Tumblr) was not perceived as sarcastic by the majority of the annotators, which highlights the existing gap between the authors' interpretation of sarcasm and that of the readers, and the amount of noise we can expect in the dataset. 
In summary, the majority of the annotators found that both the text and the image are necessary to correctly evaluate the tone of the post in more than one third of the examples (37.40\% for Instagram, 44.50\% for Tumblr). Among these, 51.07\% of the Instagram posts and 44.27\% of the Tumblr posts were agreed to be sarcastic by at least 80\% of the annotators (D-80), and 22.99\% (IG) and 31.69\% (TU) were unanimously declared sarcastic (D-100). 

\section{Automated Methods for Sarcasm Detection}
\label{sec:methods}

We investigate two automatic methods for multimodal sarcasm detection. The first, a linear Support Vector Machine (SVM) approach, has been commonly used in prior work, though this prior work has relied on features extracted mainly from the text of the post (or set of posts). In our proposal, we combine a number of NLP features with visual features extracted from the image. The second approach relies on deep learning to fuse a deep network based representation of the image with unigrams as textual input. For both of these approaches, we evaluate the individual contributions of the respective textual and visual features, along with their fusion, in Section~\ref{sec:evaluation}.

\subsection{SVM Approach}
\label{sec:svm}

For all experiments within this approach, we train a binary classification model using the sklearn toolkit\footnote{\url{http://scikit-learn.org/}} with its default settings.\footnote{We acknowledge that performance could be improved by experimenting with different parameters and kernels, however, our focus is not on optimizing for the best sarcasm detection system, but rather to construct a framework with which to show that visual features can complement textual features.}

\vspace{0.2cm}
\noindent \textbf{NLP Features.}  Our goal here is to replicate the prior art in developing a strong baseline composed of NLP features from which to investigate the impact that images have in detecting sarcasm.  
We adopt features commonly found in the literature: lexical features which measure aspects of word usage and frequency, features which measure the sentiment and subjectivity of the post, and word sequences (n-grams). We also make use of word embeddings, which has seen limited application to this task, save for a few works, such as \cite{ghosh-guo-muresan:2015:EMNLP}, but has been used as a strong baseline in the sister task of sentiment analysis \cite{faruqui-dyer:2015:ACL-IJCNLP}.  Finally, we select some of our best performing features and create a combination feature class. A description of each class is listed below:

\begin{itemize}
\item {\bf lexical}: average word length, average word log-frequency according to the Google 1TB N-gram corpus,\footnote{\url{https://catalog.ldc.upenn.edu/LDC2006T13}} number of contractions in sentence, 
average formality score as computed in~\cite{pavlick-nenkova:2015:NAACL-HLT}.
\item {\bf subjectivity}: subjectivity and sentiment scores as computed by the TextBlob module,\footnote{\url{https://textblob.readthedocs.org/en/dev/}} number of passive constructions, number of hedge words, number of first person pronouns, number of third person pronouns.
\item {\bf n-grams}: unigrams and bigrams represented as one-hot features.
\item {\bf word2vec}: average of word vectors using pre-trained word2vec embeddings \cite{mikolov2013distributed}. OOV words are skipped.
\item {\bf combination}: n-grams, word2vec and readability features (these include length of post in words and characters, as well as the Flesch-Kincaid Grade level score~\cite{kincaid1975derivation}).
\end{itemize}

Text is tokenized using nltk.\footnote{\url{http://www.nltk.org/}} In addition, we treat hashtags in Instagram and Twitter, and tags in Tumblr, as well as emojis, as part of the text on which the features are derived from.

\vspace{0.2cm}
\noindent \textbf{Visual Semantics Features (VSF).} A key module to detect sarcasm is to understand the semantics in images. We employ the visual semantics models from Yahoo Flickr Creative Commons 100M (YFCC100M)~\cite{thomee2016yfcc100m}, which include a diverse collection of complex real-world scenes, ranging from 200,000 street-life-blogged photos by photographer Andy Nystrom to snapshots of daily life, holidays, and events. Specifically, the semantics models were built with an off-the-shelf deep convolutional neural network using the Caffe framework~\cite{jia2014caffe}, and the penultimate layer of the convolutional neural network output as the image-feature representation for training classifiers for 1,570 concepts which are popular in YFCC100M. Each concept classifier is a binary support vector machine, for which positive examples were manually labeled based on targeted search/group results, while the negatives drew negative examples from a general pool. The classifiers cover a diverse collection of visual semantics in social media, such as people, animals, objects, foods, architecture, and scenery, and will provide a good representation of image contents. Examples of concepts include terms such as ``head'', ``nsfw'', ``outside'', and ``monochrome''. In our experiments, we use the output of the content classifiers as one-hot features for the SVM regression model. Essentially, if a concept is detected, no matter what its associated confidence score, we treat it as a one-hot feature.

\vspace{0.2cm}
\noindent \textbf{Multimodal Fusion.} We concatenate the textual and visual features into a long vector, and once again use the linear SVM to train the fusion model. Previous research suggests that linear SVMs are fit for text classification~\cite{Joachims:1998:TCS:645326.649721}, and our experiments find that linear SVM works very robustly to combine different kinds of features.

\subsection{Deep Learning Approach}
\label{sec:dl}

\begin{figure}[t]
\centering
\includegraphics[width=0.8\columnwidth]{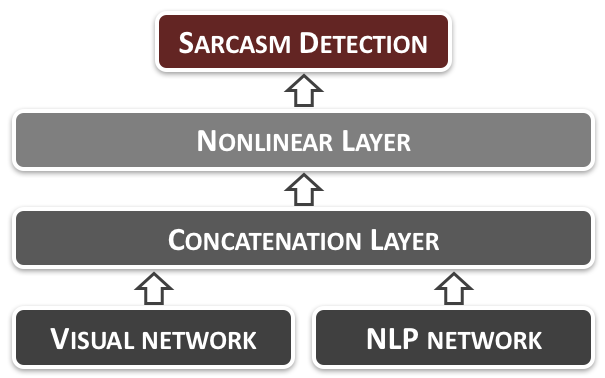}
\caption{Network structure of our model. The visual network in the figure is initialized with the model weights in \cite{DBLP:conf/bmvc/Chatfield/devil} trained on ImageNet.}
\label{fig-adaptation-network}
\end{figure}

\noindent \textbf{Adapted Visual Representation (AVR).} The visual semantics classifiers described in the previous section are limited by a fixed vocabulary. To get a stronger visual representation, we follow the work in \cite{cnnfeature-off-shelf-cvprw-14} and \cite{flickr-style-bmvc2014} that adopt a deep neural network. We borrow a model trained on ImageNet exactly from~\cite{DBLP:conf/bmvc/Chatfield/devil}, which is based on roughly one million images annotated with 1,000 object classes. There are originally seven layers in the model, but we remove the last layer of 1,000 neurons which correspond to the objects in ImageNet. The second to last layer has 4,096 neurons, which we will use to fine-tune with sarcastic and non-sarcastic data.

\vspace{0.2cm}
\noindent \textbf{Textual Features.} 
If we were to use all the NLP features in Section~\ref{sec:svm}, our deep learning framework would quickly overfit given the limited size of the training set. As a consequence, a subset of the textual features were used in this fusion method. The NLP network is a two 
two layer perceptron 
based on unigrams only. The size of the first layer of the NLP network is the size of the unigram vocabulary for every platform. We employ a hidden layer in the NLP network with 512 hidden neurons, which is comparable with the number of neurons in the AVR.

\vspace{0.2cm}
\noindent \textbf{Multimodal Fusion via Deep Network Adaptation.} 
Figure~\ref{fig-adaptation-network} illustrates the neural network adaptation framework.   
We initialize a network with fixed image filters from the ImageNet model and random weights in other layers, and adapt it to our data. This adaption framework works with the deep CNN trained on ImageNet. The concatenation layer has 4,608 neurons.
We use the rectify function as the activation function on all the nonlinear layers except for the last layer, which uses softmax over the two classes (sarcastic vs. non-sarcastic). Since in practice it is hard to find the global minimum in a deep neural network, we use Nesterov Stochastic Gradient Decent with a small random batch (size = 128). We finish training after 30 epochs.

\section{Evaluation}
\label{sec:evaluation}

We evaluate our two methods under the same conditions, and with two different evaluation settings. For the first evaluation, models are developed on the data as described in Section~\ref{sec:data}, where we train on 50\% of the data and evaluate on the remaining 50\%. Please recall that the three data sets are evenly split between sarcastic and non-sarcastic posts, with the Instagram and Tumblr data sets containing a total of 20K posts each, and Twitter totaling 4,050 posts.
We call this the {\bf Silver Evaluation}, since the data is dependent on the authors correctly labeling their posts as sarcastic. As we saw in Table~\ref{tab:cfdistrib}, 24.8\% and 31.8\% of Instagram and Tumblr posts marked by the authors as sarcastic are actually not sarcastic. For both the SVM and deep learning methods, we show results for \textit{Text-Only}, \textit{Image-Only} and the fusion of both modalities.  

Next, we evaluate the respective Instagram and Tumblr models on the crowd-curated data sets in Section~\ref{sec:crowdsourcing} (henceforth {\bf Gold Evaluation}). Unlike the evaluation on the silver sets, the models are tested on re-judged data, and thus are of much higher quality, though there are fewer examples.  

We use accuracy as our evaluation metric, and the baseline accuracy is 50\% since all sets are evenly split.

\subsection{Fusion with SVM}

\subsubsection{Evaluation on Silver Set}

\begin{table}[ht!]
\centering
\begin{tabular}{|c|ccc|}
\hline
\textbf{Feature Set} & \textbf{IG} & \textbf{TU} & \textbf{TW} \\
\hline
lexical 		& 56.7 & 54.3 & 57.8 \\
subjectivity 	& 61.7 & 59.9 & 58.3 \\
1,2-grams 		& 80.7 & 80.0 & 78.6 \\
word2vec 		& 74.9 & 73.6 & 75.3 \\
combination 	& 81.4 & 80.9 & {\bf 80.5} \\
\hline
\hline
VSF only   & 68.8 & 65.7 & 61.7        \\
\hline
\hline
n-gram $+$ VSF & 81.7 & 80.6 & 79.0 \\
combination $+$ VSF & {\bf 82.3} & {\bf 81.0}   & 80.0 \\
\hline
\end{tabular}
\caption{Silver Set evaluation using SVM fusion.}
\label{tbl:nlp-results}
\end{table}

We first evaluate the contribution of the individual NLP features from Section~\ref{sec:svm} on the three data sets, as shown in the first main block in Table~\ref{tbl:nlp-results}. The top individual feature is n-gram (1- and 2-grams), roughly performing at close to 80\% accuracy across all data sets. In fact, even though we use three disparate data sets, the performance figures for each feature are consistently the same as the the ranking of the features. This may suggest that users do not alter the way they use sarcasm across platforms, though the best way of testing this hypothesis would be to investigate whether models trained on one platform, e.g., Twitter, can approximate the performance found on the other platforms, e.g., Instagram, when models are trained on native data. Finally, merging several of the feature classes into one (\textit{combination}) yields the best performance, exceeding 80\% for all data sets.

Using only the visual semantics features (VSF) yields an accuracy around 65\% across the data sets. This is more than 15 points lower than the best NLP models; however, we were surprised that such a simple feature class actually outperformed the lexical and subjectivity features, both of which have been used in prior NLP work for the sarcasm detection task.  

Finally, we combine the visual semantics features with the two best performing NLP features, i.e., n-grams and the combination feature class (last two rows of Table~\ref{tbl:nlp-results}). For all the three data sets, the model with n-grams + VFS outperformed the model solely trained on n-grams by a small margin. However, it was not better than using the combination features. When combining the visual features with the combination features, we achieve the highest performance in Instagram (82.3\%) and Tumblr (81.0\%). In Twitter, the fusion produces the second highest performance (80.0\%) to the 80.5\% yielded by combination features only. These results show that including simple, noisy image-related features can improve sarcasm detection, albeit by a small margin.  

\subsubsection{Evaluation on Gold Set}

Next, we investigate how well our models perform on the curated gold sets in Instagram and Tumblr. For the sake of simplicity, we focus our NLP evaluation on just the two top performing feature classes: n-grams and combination.

\begin{table}[htbp]
\centering
\begin{tabular}{|c|ccc|}
\hline
\multirow{2}{*}{\textbf{Feature Set}} & \textbf{D-50} & \textbf{D-80} & \textbf{D-100} \\
 & {\small \emph N$=$374} & {\small \emph N$=$191} & {\small \emph N$=$86} \\
 \hline
1,2-grams		& 81.7 & 81.9 & 80.2 \\
combination 	& 81.7 & 82.5 & 80.2  \\
\hline
\hline
VSF only   & 75.7 & 72.8 & 68.0      \\
\hline
\hline
1,2-grams $+$ VSF & {\bf 86.6} & {\bf 87.7} & {\bf 83.7}  \\
comb. $+$ VSF  & 84.8 & 85.3 & 80.8 \\
\hline
\end{tabular}
\caption{SVM evaluation on Instagram Gold Sets.}
\label{tbl:instagram-gold-results}
\end{table}

\begin{table}[htbp]
\centering
\begin{tabular}{|c|ccc|}
\hline
\multirow{2}{*}{\textbf{Feature Set}} & \textbf{D-50} & \textbf{D-80} & \textbf{D-100} \\
 & {\small \emph N$=$445} & {\small \emph N$=$197} & {\small \emph N$=$141} \\
\hline
1,2-grams		& 88.3 & 84.8 & 84.0 \\
combination 	& {\bf 88.8} & 86.0 & 84.4 \\
\hline
\hline
VSF only   & 70.7 & 73.1 & 73.8       \\
\hline
\hline
1,2-grams $+$ VSF & 88.5 & {\bf 87.8} & {\bf 89.7} \\
comb. $+$ VSF & 88.0 & 87.1 & {\bf 89.7} \\
\hline
\end{tabular}
\caption{SVM evaluation on Tumblr Gold Sets.}
\label{tbl:tumblr-gold-results}
\end{table}

Table~\ref{tbl:instagram-gold-results} shows the results for the different modalities in Instagram. For the NLP features, the combination and n-gram are tied for the 50\% and 100\% agreement conditions (D-50 and D-100), while combination narrowly outperforms its counterpart in the 80\% condition (D-80). As in the previous silver results, using the VSF only causes a loss in performance of nearly 15 points. The best results come from fusing n-grams with VSF, yielding a performance improvement of about 5\% on all three agreement levels. Interestingly, while combination $+$ VSF was generally the best feature in the silver evaluation, it is the second best here.

The Gold Tumblr results in Table \ref{tbl:tumblr-gold-results} show a similar pattern with Table \ref{tbl:instagram-gold-results}: the combination features outperform the n-gram features by a small margin across all three agreement levels, and only using VSF results in a performance loss of around 15 points accuracy compared to combination. We see the best performance when fusing the NLP and VSF features. At the 80\% agreement level, n-gram $+$ VSF yields a performance of 87.8\%, which outperforms the best non-fusion performance by 1.8 points (86.0\%). At the 100\% agreement level, both fusion sets perform at 89.7\%, a 5\% point improvement. However, at the lower agreement rate (50\%), the best performing fusion method just narrowly misses the combination method (88.5\% to 88.8\%).

The main message from both the silver and gold evaluations is that incorporating simple features which describe the image in a very traditional framework can improve performance. In general, the best performance comes not from fusing VSF with combination features, but rather with n-grams. We speculate that this may be due to the mismatch between the silver and gold sets. We do note that in some cases the performance improvement was small or non-existent. This is partially due to the noisiness of the data, the high baseline set by the NLP features, and also the accuracy of the VSF features, which can be viewed as hypotheses of what the classifier believes is present in the photo, even if weakly present.

\subsection{Fusion with Deep Network Adaptation}

Next, we evaluate our deep learning approach on our silver and gold sets. We additionally evaluate the model with image (AVR) and text (unigram) features only, for which the concatenation layer (see Figure~\ref{fig-adaptation-network}) still exists but has no effect with single modality input. The three models use the same learning rates.

\subsubsection{Evaluation on Silver Set}

\begin{table}[htbp]
\centering
\begin{tabular}{|c|ccc|}
\hline
\textbf{Feature Set} & \textbf{IG} & \textbf{TU} & \textbf{TW} \\
\hline
1-grams & 71.0 & 65.3 & 54.1 \\
AVR only & 73.8 & 69.2 & 68.7 \\
\hline
1-grams $+$ AVR  & {\bf 74.2} & {\bf 70.9} & {\bf 69.7} \\
\hline
\end{tabular}
\caption{Silver Set evaluation using DNA fusion.}
\label{tbl:dl-silver results}
\end{table}

Table \ref{tbl:dl-silver results} shows the the evaluation on the silver set. It is easy to see that fusing the textual and image signals together provides the best performance across all three sets, ranging from 74.2\% in Instagram to 69.7\% in Twitter. That confirms our hypothesis that the visual aspect plays a role in the detection of sarcasm.

Another interesting phenomenon is that the image-only network outperforms the visual semantics features consistently in all three platforms: 73.8\% vs. 68.8\% in Instagram, 69.2\% vs. 65.7\% in Tumblr, and 68.7\% vs. 61.7\% in Twitter. This suggests that the adapted deep CNN better captures the diversity of sarcastic images. On the other hand, our text-based network is worse than the text models using SVM. The reason is mainly because our text network does not use bigrams or higher dimensional features. Since the visual semantics features are not fine-tuned, the simpler fusion by SVM method does not overfit the training set. As a result, all state-of-the-art NLP features described in Section~\ref{sec:svm} can be used in this model.

Among the three platforms, the performance in Twitter is lower than in the other two. We believe that this is mainly due to the small amount of training data (2,000 posts), which is an issue for deep learning. Also, given that Twitter is mostly a textual platform (especially compared to the more image-centric Instagram and Tumblr), the weaker textual baseline seems to fail to capture the nuances of sarcasm used in this platform.

\subsubsection{Evaluation on Gold Set}

\begin{table}[htbp]
\centering
\begin{tabular}{|c|ccc|}
\hline
\multirow{2}{*}{\textbf{Feature Set}} & \textbf{D-50} & \textbf{D-80} & \textbf{D-100} \\
 & {\small \emph N$=$374} & {\small \emph N$=$191} & {\small \emph N$=$86} \\
\hline
1-grams 		& 69.7 & 67.7 & 63.1 \\
AVR only 	& 77.0 & 74.6 & 74.8 \\
\hline 
1-grams $+$ AVR   &\textbf{77.8} &\textbf{78.4} & \textbf{77.6}   \\
\hline
\end{tabular}
\caption{DNA evaluation on Instagram Gold Sets.}
\label{dl-gold-instagram}
\end{table}

\begin{table}[htbp]
\centering
\begin{tabular}{|c|ccc|}
\hline
\multirow{2}{*}{\textbf{Feature Set}} & \textbf{D-50} & \textbf{D-80} & \textbf{D-100} \\
 & {\small \emph N$=$445} & {\small \emph N$=$197} & {\small \emph N$=$141} \\
\hline
1-grams 			& 68.4 & 65.8 & 64.6 \\
AVR only 	& 75.8 & 74.6 & {\bf 75.5} \\
\hline 
1-grams $+$ AVR  & {\bf 77.6} & {\bf 75.6} & 74.7       \\
\hline
\end{tabular}
\caption{DNA evaluation on Tumblr Gold Sets.}
\label{dl-gold-tumblr}
\end{table}

Our gold results show a similar pattern. In the Tumblr set, the fusion of text and image yields the best performance over D-50 and D-80, but is narrowly behind just using the image on D-100. In the Instagram set, the fusion of text and images yields the best performance in all three platforms. Since the text feature is limited, the performance of deep network adaptation is not as competitive as the SVM based fusion method. However, we think the performance of deep neural network adaption will be improved with more training examples.

\section{Conclusions}
\label{sec:conclusions}

To the best of our knowledge, this work represents the first empirical investigation on the impact of images for sarcasm detection in social media. In particular, we first investigate the role of images, and quantitatively show that humans use visuals as situational context to decode the sarcastic tone of a post. The collected and annotated data will be shared with the research community. Second, we show that automatic methods for sarcasm detection can be improved by taking visual information into account. Finally, while most previous work has focused on the study of textual utterances on Twitter, our research shows breadth by tackling two other popular social media platforms: Instagram and Tumblr.

We propose two types of multimodal fusion frameworks to integrate the visual and textual components, and we evaluate them across three social media platforms with heterogeneous characteristics. With the use of visual semantics features, we observe an improved performance for the noisy dataset in the case of Instagram (the most image-centric platform), while the impact of images in Tumblr and Twitter was not perceived as relevant. We argue that this behavior is due to their text-centric nature. In the case of curated data though, we observe higher predictive accuracy across all the platforms, and across almost all of the agreement levels, which suggests that the visual component plays an important role when human judgments are involved.  

By using deep network adaptation, we show a consistent increment in performance across the three platforms. Also in this case, Instagram was the platform that reached the highest accuracy. We have pointed out the weak performance of the textual features used in the deep learning approach. The challenges that prevent us from using more advanced textual features (such as those used in the SVM model) are two-fold: 1) given the limited size of the training set, the network adaptation method suffers from overfitting; adding new features does not help when the fusion network can get almost perfect accuracy on the training set; and 2) a higher dimensionality brings difficulties for a fast neural network training due to the limitations of the GPU memory. Collecting more training data should, at the very least, address the overfitting issue.

Images can be thought of as another form of contextual clue, much like the role of previous tweets by a user or the overall sarcasm levels of a discussion thus far. In our future work, we wish to build a model which integrates all these contextual clues within our framework to assess which ones have the largest impact per platform. We are also interested in including visual sentiment frameworks in the evaluation of the sarcastic tone.

\section{Acknowledgments}
This work is partially supported by the project ``ExceptionOWL: Nonmonotonic Extensions of Description Logics and OWL for defeasible inheritance with exceptions'', Progetti di Ateneo Universit\`a  degli Studi di Torino and Compagnia di San Paolo, call 2014, line ``Excellent (young) PI''.

%

\bibliographystyle{abbrv}
\bibliography{bib/sarcasm}  %

\end{document}